\documentclass{article}
\usepackage{spconf,amsmath,graphicx}
\usepackage{amsfonts}
\usepackage{multirow}
\usepackage{array}
\usepackage[switch]{lineno}

\usepackage{enumitem}
\setlist{nosep, leftmargin=14pt}

\usepackage{mwe} 

\title{Semantic-aware Temporal Channel-wise Attention for Cardiac Function Assessment}
\name{Guanqi Chen \qquad Guanbin Li\sthanks{Corresponding author.}}
\address{School of Computer Science and Engineering, Sun Yat-Sen University, China}
\begin{document}
%
\maketitle
\begin{abstract}
Cardiac function assessment aims at predicting left ventricular ejection fraction (LVEF) given an echocardiogram video, which requests models to focus on the changes in the left ventricle during the cardiac cycle. How to assess cardiac function accurately and automatically from an echocardiogram video is a valuable topic in intelligent assisted healthcare. Existing video-based methods do not pay much attention to the left ventricular region, nor the left ventricular changes caused by motion. In this work, we propose a semi-supervised auxiliary learning paradigm with a left ventricular segmentation task, which contributes to the representation learning for the left ventricular region. To better model the importance of motion information, we introduce a temporal channel-wise attention (TCA) module to excite those channels used to describe motion. Furthermore, we reform the TCA module with semantic perception by taking the segmentation map of the left ventricle as input to focus on the motion patterns of the left ventricle. Finally, to reduce the difficulty of direct LVEF regression, we utilize an anchor-based classification and regression method to predict LVEF. Our approach achieves state-of-the-art performance on the Stanford dataset with an improvement of 0.22 MAE, 0.26 RMSE, and 1.9\% $R^2$. 
\end{abstract}
\section{Introduction}
\label{sec:intro}

Cardiac function is the capability of the heart to meet the metabolic demands of the body, which has a significant impact on human health. If the cardiac function is slightly damaged, fatigue, palpitations, dyspnea, or angina pectoris may occur, and in severe cases, heart failure may occur. In recent years, there are more patients with cardiac dysfunction than ever before, which leads to heart dysfunction becoming a global health problem \cite{ziaeian2016epidemiology}. Therefore it is very important to timely detect and treat cardiac dysfunction, which relies on an accurate assessment of cardiac function. \par
Left ventricular ejection fraction (LVEF), the ratio of the stroke volume to the end-diastolic volume in the left ventricle, is a critical metric of cardiac function. It reveals the effectiveness of pumping into the systemic circulation. In recent years, it has aroused great interest to use deep learning techniques \cite{ghorbani2020deep,ouyang2020video} on echocardiography to estimate LVEF.
\cite{ghorbani2020deep} tried to use a 2D CNN to assess cardiac function based on the manually selected images at end-systole and end-diastole. However, this simple method had a substantial error compared to the human assessment of cardiac function. 
\cite{ouyang2020video} attempted to predict LVEF based on the echocardiogram video.
They employed a spatiotemporal convolutional network to directly regress LVEF. Then they utilized another convolutional neural network (CNN) to segment left ventricles, which was used to identify the cardiac cycles. And they aggregated the results of several cardiac cycles to obtain the final prediction. Compared with \cite{ghorbani2020deep}, the method proposed in \cite{ouyang2020video} made a great progress, which might benefit from the motion information in the echocardiogram video. However, current video-based methods ignore the positive effect of the left ventricular segmentation task on the cardiac function assessment task. The cardiac function assessment task requires models to focus on the changes of the left ventricle during the cardiac cycle, while the segmentation task can make networks capture discriminative representation for the left ventricular region. Besides, the left ventricular masks generated by the segmentation network can serve as cues, which include the spatial information of the left ventricles, to benefit the cardiac function assessment task. \par


\begin{figure}[tbp]
\centering
\includegraphics[width=0.45\textwidth]{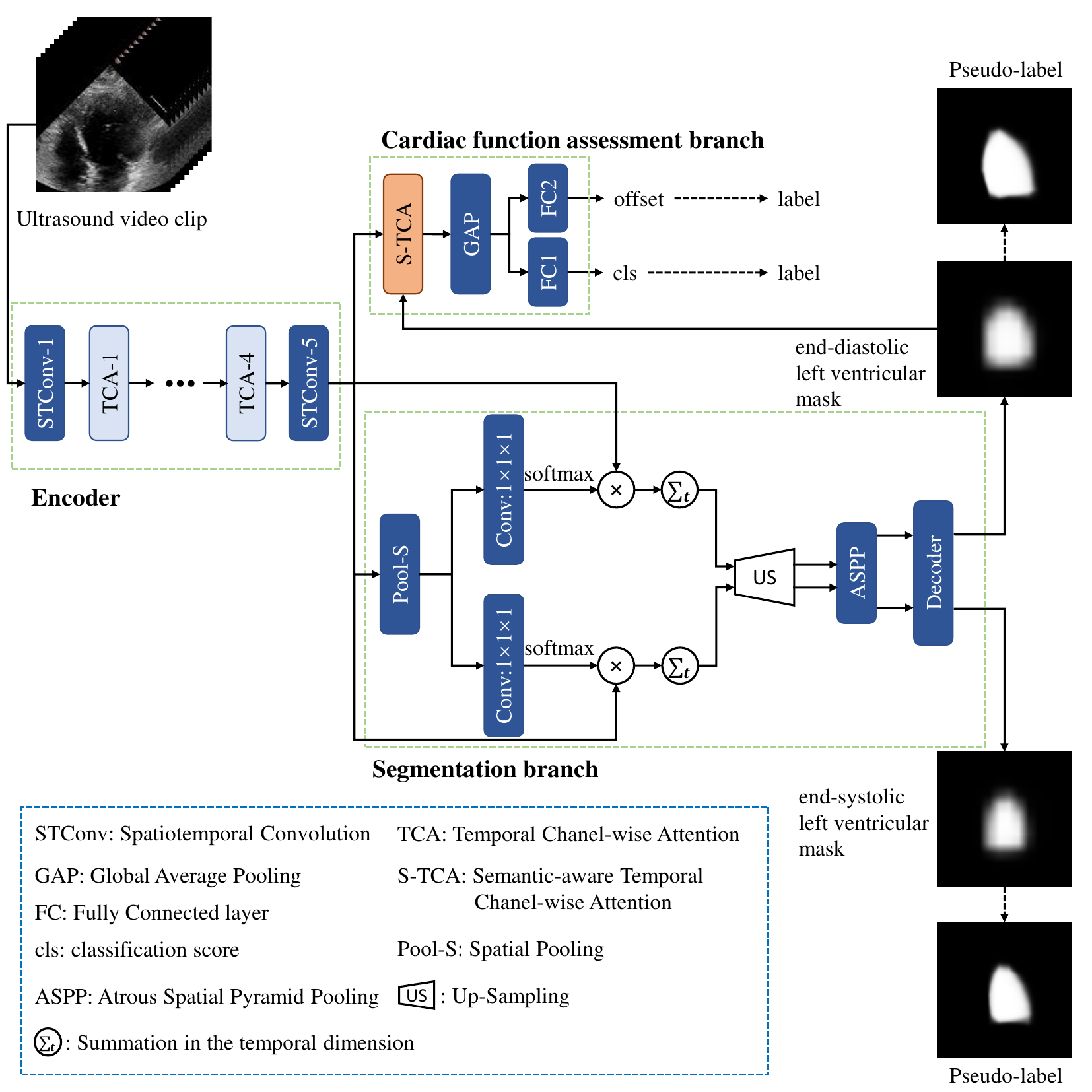}
\caption{Overview of our proposed method. Our model is composed of a shared encoder and two branches for left ventricular segmentation and cardiac function assessment respectively.} \label{fig-framework}
\end{figure}

Motivated by the above observations, in this paper, we propose a semi-supervised auxiliary learning paradigm to jointly address cardiac function assessment and left ventricular segmentation tasks. In the paradigm, we utilize a shared encoder and two different branches to predict LVEF and segment left ventricles respectively, which can make the encoder capture a better representation of the left ventricle. To make full use of the motion information, we design a temporal channel-wise attention (TCA) module to excite those channels which are used to describe motion in the spatiotemporal features. Besides, with the help of left ventricular segmentation masks, we further reform the TCA module with semantic perception, which transforms the TCA module into semantic-aware temporal channel-wise attention (S-TCA) module. Through the S-TCA module, we can make our model focus on the motion patterns of the left ventricle. Moreover, inspired by anchor-based detection methods \cite{girshick2015fast}, we introduce an anchor-based classification and regression method to predict LVEF instead of performing direct LVEF regression. In this way, we reduce the variance of the training samples, which is easier for the network to learn the potential expression. Finally, our approach achieves state-of-the-art performance on the Stanford dataset \cite{ouyang2020video}. 

\section{Methodology}
\label{sec:method}

The overall architecture of our proposed method is shown in Figure~\ref{fig-framework}. A spatiotemporal convolution network with a set of temporal channel-wise attention (TCA) modules is employed to obtain the feature representation for the input video clip. And these TCA modules are designed to excite the motion patterns at a frame-level. After that, two branches are applied to address left ventricular segmentation and cardiac function assessment tasks, respectively. With the predicted left ventricular segmentation masks, we use a semantic-aware temporal channel-wise attention (S-TCA) module to bridge two branches and excite the motion patterns of the left ventricle. Finally, we utilize an anchor-based classification and regression method to predict LVEF. In this section, we will elaborate our proposed components.


\subsection{Temporal Channel-wise Attention}
Spatiotemporal features captured by spatiotemporal convolutions are commonly composed of static appearance features and dynamic motion features. Due to the high frame rate and the fixed view of echocardiography, the appearance features of adjacent frames are similar. However, the motion patterns of the adjacent frames may be different, especially at end-systole and end-diastole. The direction of motions in these periods is the opposite of that in previous frames. Motivated by this observation, we argue that the appearance features after the max-pooling operation and the mean-pooling operation in the adjacent frames are the same, while the motion features may be different. Then we can leverage this difference to excite the motion features through a group of convolutions and activation functions. \par
\begin{figure}[tbp]
\centering
\includegraphics[width=0.45\textwidth]{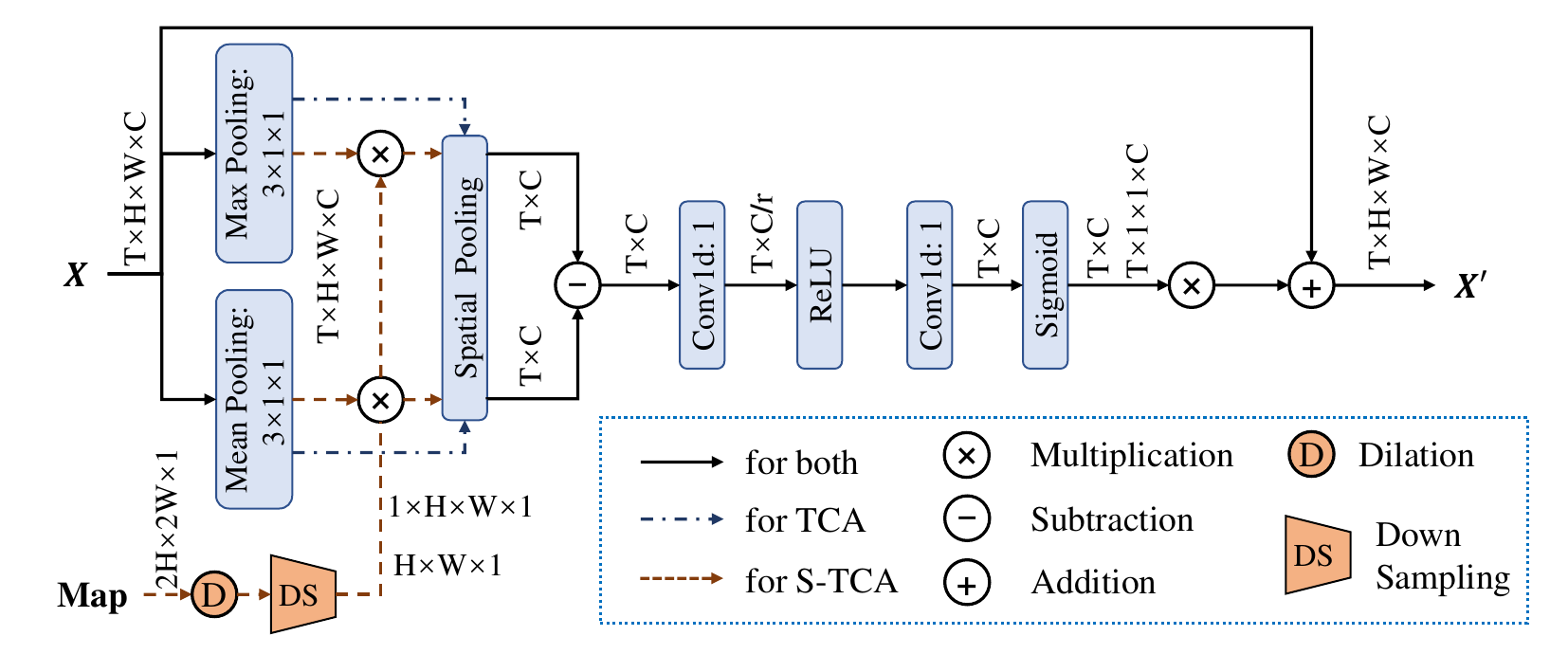}
\caption{Illustration for temporal channel-wise attention (TCA) module and semantic-aware temporal channel-wise attention (S-TCA) module.} \label{fig-tca_s_tca}
\end{figure}

The architecture of our proposed TCA module is illustrated in Figure~\ref{fig-tca_s_tca}. Given an input feature $X \in \mathbb{R}^{T \times H \times W \times C}$, local max-pooling and mean-pooling are applied to aggregate the information from adjacent frames and obtain features $X^{max}$, $X^{mean}$ $\in \mathbb{R}^{T \times H \times W \times C}$. 
Then a global average pooling in the spatial dimension is adopted to $X^{max}$ and $X^{mean}$ to obtain global representations $\bar{X}^{max}$, $\bar{X}^{mean}$ $\in \mathbb{R}^{T \times C}$.
After that, based on the above analysis, we directly apply subtraction to global representations to stress motion features.  
Inspired by the SE block \cite{hu2018squeeze}, we utilize two 1D convolutions with ReLU and sigmoid activation functions to obtain the temporal channel-wise attention weights $E \in \mathbb{R}^{T \times C}$. 
\begin{equation}
E = sigmoid(W_2*(ReLU(W_1*(\bar{X}^{max} - \bar{X}^{mean}))),
\end{equation}
where $*$ denotes the convolution operation. $W_1 \in \mathbb{R}^{C/r \times C \times 1}$ and $W_2 \in \mathbb{R}^{C \times C/r \times 1}$ are the parameters of 1D convolutions. $r = 16$ is the reduction ratio.
Finally, we leverage $E$ to excite the motion patterns of $X$ in the following process:
\begin{equation}
X' = X \cdot E + X.
\end{equation}
In this way, we can highlight the motion features without discarding the appearance features.

\subsection{Semi-supervised Auxiliary Learning Paradigm}
In order to make our model focus on the left ventricle, we introduce an auxiliary task to segment end-diastolic and end-systolic left ventricles. However, each video has only one annotated cardiac cycle with two masks for the end-diastolic and end-systolic left ventricle, which can not ensure that there exist ground-truths for randomly sampled video clips. 
To address this issue, we utilize a DeeplabV3 \cite{chen2017rethinking} model trained on the sparse labels to generate the left ventricular masks for those unlabeled frames. Then we choose the predicted masks with the largest and smallest area among the input video clips as pseudo-labels to train the segmentation branch. \par
As shown in Figure~\ref{fig-framework}, the segmentation branch contains two paths to segment the end-diastolic and end-systolic left ventricular masks. Next, we take how to obtain the end-diastolic left ventricular mask as an example to elaborate. Firstly, a spatial pooling is applied to the spatiotemporal feature $Z\in \mathbb{R}^{T \times H \times W \times C}$ to get the global representation $\bar{Z} \in \mathbb{R}^{T \times C}$. Then a 1D convolution and a softmax activation function are employed to $\bar{Z}$ to obtain the adaptive weight $R \in \mathbb{R}^{T \times 1}$ which represents the temporal relevance to the end-diastole. 
\begin{equation}
R = softmax(W_3 * \bar{Z}),
\end{equation}
where $W_3 \in \mathbb{R}^{1 \times C \times 1}$ is the parameter of 1D convolution. After that, we obtain the end-diastolic representation $F \in \mathbb{R}^{H \times W \times C}$ through the following operation:
\begin{equation}
F = \sum_{t=1}^{T} Z_t \cdot R_t.
\end{equation}
Then we upsample the representation $F$ for higher resolution, which is the same as the feature size in the DeeplabV3 model before the ASPP \cite{chen2017rethinking} module. Finally, we use an ASPP module and a decoder which consists of two convolutional layers to produce the end-diastolic left ventricular mask.\par 

\subsubsection{Semantic-aware temporal channel-wise attention}
With the help of left ventricular masks, we can further reform the TCA module to focus on the motion patterns of the left ventricle. As shown in Figure~\ref{fig-tca_s_tca}, the semantic-aware temporal channel-wise attention (S-TCA) module utilizes the end-diastolic left ventricular mask to conceal the trivial region. Since sometimes the predicted mask is not accurate to cover the whole left ventricle, we expand the predicted area of left ventricle via dilation operation, which is achieved by the max-pooling operation.

\subsection{Anchor-based Classification and Regression}
Instead of directly regressing LVEF, we predict LVEF through two parts: classifying its interval and regressing the corresponding center offset.
Firstly, we set $M=20$ equally spaced anchor intervals, which cover the range of LVEF. Then we use two sibling fully connected layers following a global average pooling in spatiotemporal dimension to predict results. The first output is a vector of classification score for each anchor interval, and we can utilize a softmax function to turn it into a discrete probability distribution $p \in \mathbb{R}^{M}$. The second fully connected layer outputs a vector of offset $o \in \mathbb{R}^{M}$ for each anchor interval. And their corresponding labels, the ground-truth class $u \in \mathbb{R}^{1}$ and the ground-truth offset $v \in \mathbb{R}^{M}$, are obtained by following:
\begin{equation}
\begin{split}
& u = \lfloor y/l \rfloor, \\
& v_m = \frac{y - c_m}{l},
\end{split}
\end{equation}
where $y$ is the ground-truth of LVEF, and $l$ is the size of the anchor interval, which is equal to $100/M$. $m$ denotes the index of anchor intervals, and $c_m$ is the median value of the m-th anchor interval. 


In this work, we utilize a multi-task loss function to train the whole model:
\begin{equation}
L = L_{ef} + \beta L_{aux},
\end{equation} 
where $L_{ef}$ and $L_{aux}$ represent the loss of the cardiac function assessment task and the auxiliary task. $L_{ef}$ is consist of two parts, the cross-entropy loss of the anchor interval and the smooth $L_1$ loss \cite{girshick2015fast} of the offset. And $L_{aux}$ is the cross-entropy loss between pseudo-labels and predicted left ventricular masks. $\beta$ is a hyperparameter for balancing two loss terms, which is set to 0.01 in our experiments since $L_{aux}$ is two orders of magnitude larger than $L_{ef}$. 

\section{Experiments}
In this section, we evaluate our proposed components on the Stanford dataset \cite{ouyang2020video}. It is composed of 10030 labeled echocardiogram videos, including 7465 videos for training, 1288 videos for validation, and 1277 videos for testing. All videos are apical-4-chamber view, and they consist of a series of gray-scale images of $112 \times 112$ pixels.
%


\subsection{Evaluation Metrics}

In this paper, we only focus on the performance of the cardiac function assessment task, that is, the quality of the predicted LVEF. To quantitatively measure the model performance, we adopt three popular evaluation criteria: MAE, RMSE, and $R^2$. 

\subsection{Comparison with State-of-the-art Methods}
In this section, we compare our proposed method with the existing state-of-the-art cardiac function assessment algorithms on the Stanford test set, which is shown in Table \ref{tab1}. EchoNet-Dynamic \cite{ouyang2020video} uses a R(2+1)D model \cite{tran2018closer} to regress the LVEF, and employs the beat-to-beat evaluation strategy to assess the cardiac function. For fair comparisons, we utilize its open-source code to train several 3D CNNs which include MC3 \cite{tran2018closer}, R3D \cite{tran2015learning}, and R(2+1)D\cite{tran2018closer}, in the same setting, then evaluate them with the same strategy of our model, which is to randomly sample 10 different clips from the video, and average the predictions of them to obtain the final prediction. As Table \ref{tab1} displays, the proposed method achieves the lowest MAE and RMSE, the highest $R^2$ on the Stanford test set. Compared to the result reported in \cite{ouyang2020video}, our method considerably outperforms it by 0.22 MAE, 0.26 RMSE and 1.9\% $R^2$.

\begin{table}[!t]
\centering
\caption{Comparison results with state-of-the-art algorithms on the Stanford test set. $^\star$ indicates the result is copied.}\label{tab1}
\begin{tabular}{c|ccc}
\hline
Methods& MAE$\downarrow$& RMSE$\downarrow$& $R^2\uparrow$ \\
\hline
MC3 \cite{tran2018closer}& 4.34& 5.76& 0.778\\
R3D \cite{tran2015learning}& 4.16& 5.55& 0.794\\
R(2+1)D \cite{tran2018closer}& 3.95& 5.27& 0.814\\
EchoNet-Dynamic \cite{ouyang2020video}$^\star$& 4.05& 5.32& 0.81\\
Ours& \bf 3.83& \bf 5.06& \bf 0.829\\
\hline
\end{tabular}
\end{table}

\subsection{Ablation Study}

In Table \ref{tab_ablation}, we verify the effectiveness of the components which constitute our proposed approach. $M_0$ represents the R(2+1)D model with a fully connected layer to regress LVEF directly. Based on $M_0$, $M_1$ employs the Anchor-based Classification and Regression method to predict LVEF rather than regress LVEF directly through a fully connected layer. Further, $M_2$ equips the R(2+1)D model with a set of TCA modules that follow the spatiotemporal convolution blocks. Finally, on the basis of $M_2$, $M_3$ is to make the model semantic-aware through adopting the semi-supervised auxiliary learning paradigm with the left ventricular segmentation task and replacing the last TCA module with the S-TCA module. As Table \ref{tab_ablation} indicates, each proposed component brings benefit to the complete method. 

\begin{table}[!t]
\centering
\caption{Ablation study on the Stanford test set. Sem represents semantic related strategies.}\label{tab_ablation}
\begin{tabular}{p{1.1cm}<{\centering}|p{0.5cm}<{\centering} p{0.5cm}<{\centering} p{0.5cm}<{\centering} |ccc}
\hline
Methods& ACR & TCA & Sem &  MAE$\downarrow$& RMSE$\downarrow$& $R^2\uparrow$ \\
\hline
$M_0$&  $\times$ & $\times$ & $\times$ & 3.95& 5.27& 0.814\\
$M_1$& $\checkmark$ & $\times$ & $\times$ & 3.92& 5.21& 0.819\\
$M_2$& $\checkmark$ & $\checkmark$ & $\times$ & 3.89& 5.11& 0.826\\
$M_3$& $\checkmark$ & $\checkmark$ & $\checkmark$ & \bf 3.83& \bf 5.06& \bf 0.829\\
\hline
\end{tabular}
\end{table}

\section{Conclusion}

In this paper, we propose a new method for cardiac function assessment. A semi-supervised auxiliary learning paradigm is introduced to facilitate the representation learning for the left ventricular region. Besides, TCA and S-TCA modules are designed to excite those channels used to describe motion. Finally, in order to reduce the difficulty of direct LVEF regression, we utilize an anchor-based classification and regression method to predict LVEF. Our approach achieves state-of-the-art performance on the Stanford dataset. 

\section{Compliance with Ethical Standards}
This is a numerical simulation study for which no ethical approval was required.

\section{Acknowledgments}
This work is supported by the Guangdong Basic and
Applied Basic Research Foundation under Grant No.2020B15150200 48,
the National Natural Science Foundation of China (No.619 76250), the Guangzhou Science and Technology Project (No.202102020633), and the Guangdong Provincial Key Laboratory of Big Data Computing, The Chinese University of Hong Kong, Shenzhen.

\section{Appendix}
In this supplementary material, we will elaborate more details and experimental analyses, including loss function, dataset, implementation details, discussion and experiments on proposed attention mechanism, robustness analysis, and visualization analysis.

\subsection{Loss Function}
In this work, we utilize a multi-task loss to train the whole model, which is defined as:
\begin{equation}
L = L_{ef} + \beta L_{aux},
\end{equation} 
where $L_{ef}$ and $L_{aux}$ represent the loss of the cardiac function assessment task and the auxiliary task, $\beta$ is a hyperparameter for balancing two loss terms, which is set to 0.01 in our experiments. The reason why $\beta$ is so small is that we calculate $L_{aux}$ by accumulating losses in spatial dimension instead of averaging them, which leads to that $L_{aux}$ is much greater than $L_{ef}$. 

As for the semi-supervised auxiliary task, since the quality of pseudo-labels is different, we utilize the average Dice similarity coefficient (DSC) between the sparse annotations and corresponding outputs of the DeeplabV3 model as the weight to rescale the loss for semi-supervised auxiliary learning.
\begin{equation}
\begin{split}
& L_{aux} = \alpha L_{seg} \\
& with \ \alpha = \frac{DSC(s_{ed}, {s'}_{ed}) + DSC(s_{es}, {s'}_{es})}{2}, \\
\end{split}
\end{equation}
where $ed$ and $es$ denote the end-diastole and end-systole. $s$ and $s'$ are the ground-truths of left ventricular masks and the outputs of the Deeplabv3 model respectively. $\alpha$ is the weight that represents the quality of pseudo-labels for each training video. $L_{seg}$ is the sum of the cross-entropy losses between the pseudo-labels and corresponding predictions of the segmentation branch at the end-diastole and end-systole. 

\begin{equation}
L_{aux} = 0.5\left(L_{BCE}(P_{ed}, G_{ed}) + L_{BCE}(P_{es}, G_{es})\right)
\end{equation}

As for the cardiac function assessment task, its loss $L_{ef}$ is consist of two parts, classification loss $L_{cls}$ and regression loss $L_{reg}$.
\begin{equation}
L_{ef} = L_{cls}(p,u) + L_{reg}(o,v,p),
\end{equation}
in which 
\begin{equation}
L_{cls}(p,u) = -\log p_u,
\end{equation}
\begin{equation}\label{eq_m}
L_{reg}(o,v,p) = \sum_{m=1}^{M} p_m Smooth L_1(o_m-v_m),
\end{equation}
where $p \in \mathbb{R}^{M}$ is the predicted probability (after Softmax function) for classification, $u \in \mathbb{R}^{1}$ is the ground-truth class, $o \in \mathbb{R}^{M}$ is the predicted offset vector, and $v \in \mathbb{R}^{M}$ is the ground-truth offset. $M$ is the number of anchor intervals, and $m$ is the index to enumerate the number of anchor intervals, $Smooth L_1$ represents the the smooth $L_1$ loss \cite{girshick2015fast}.

\subsection{Dataset}
The Stanford dataset \cite{ouyang2020video} is the only publicly available dataset for the video-based cardiac function assessment task. It contains a total of 10030 labeled echocardiogram videos. For each video, it contains diverse labels, including left ventricular ejection fraction (LVEF), end-systolic and end-diastolic volumes, the trace of left ventricle at end-systole and end-diastole. All of those videos are apical-4-chamber view, and they consist of a series of gray-scale images of $112 \times 112$ pixels.

\subsection{Implementation Details}
Following \cite{ouyang2020video}, an input video clip of 32 frames is generated with a sampling rate of 1 in 2 frames. As for the model architecture, a R(2+1)D model \cite{tran2018closer} with proposed temporal channel-wise attention (TCA) modules is used to capture spatiotemporal representation for the input video clip. During testing, we follow the mainstream strategy in the video classification field \cite{tran2018closer,li2020tea}, which is to randomly sample 10 different clips from the video, and average the predictions of them to obtain the final prediction.

\subsection{Discussion and Experiments on Proposed Attention Mechanism}
In this part, we want to emphasize the differences among our proposed TCA module and two representative excitation mechanisms (SENet \cite{hu2018squeeze} and TEANet \cite{li2020tea}). Compared with the SE block, our module considers the temporal information and excites those channels used to describe motion instead of the informative channels. Besides, our TCA module utilizes the residual connection to preserve the non-excited features, while the SE block ignores those non-excited features. In contrast to the ME module, the input of our module consists of motion features obtained by spatiotemporal convolutions, while the ME module needs to construct the motion features by subtracting the features of the adjacent frames. And the motivation of our TCA module is that the appearance of adjacent frames is similar and the motion may change, which is different from the ME module.

\begin{table}[!t]
\centering
\caption{Comparison results with different attention mechanisms on the Stanford test set.}\label{taba}
\begin{tabular}{c|ccc}
\hline
Methods& MAE$\downarrow$& RMSE$\downarrow$& $R^2\uparrow$ \\
\hline
SE& 3.95& 5.29& 0.813\\
ME& 3.98& 5.32& 0.811\\
TCA& \bf 3.89& \bf 5.11& \bf 0.826\\
\hline
\end{tabular}
\end{table}

\begin{table}[!t]
\centering
\caption{   }\label{taba}
\begin{tabular}{ccccc}
\hline
 & Train& Val& Test & Total \\
\hline
Num. of Videos& 7465 & 1288 & 1277 & 10030\\
\hline
\end{tabular}
\end{table}

In addition, we conduct experiments to compare the TCA module with those two attention mechanisms mentioned above. For fair comparisons, all the attention mechanisms are added to the baseline \cite{tran2018closer} with the anchor-based classification and regression method. And other experimental settings remain the same. As shown in Table~\ref{taba}, our proposed TCA module achieves the best performance, which witnesses the effectiveness of our TCA module.

\begin{figure*}[tbp]
\centering
\includegraphics[width=1.0\textwidth]{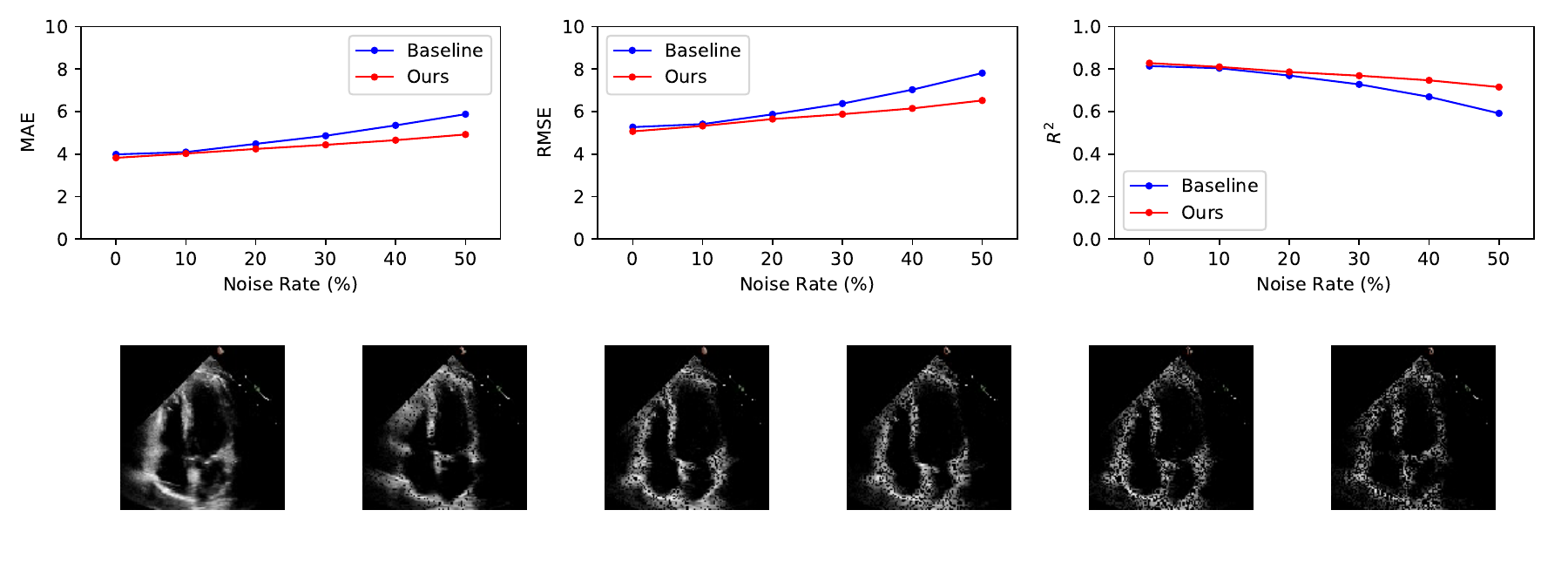}
\caption{The performance of our proposed method with different levels of noise. The images in the last row are the degraded images at six different degrees, including 0\%, 10\%, 20\%, 30\%, 40\%, 50\% noise rates.}\label{noise}
\end{figure*}

\begin{figure*}[h]
\centering
\includegraphics[width=1.0\textwidth]{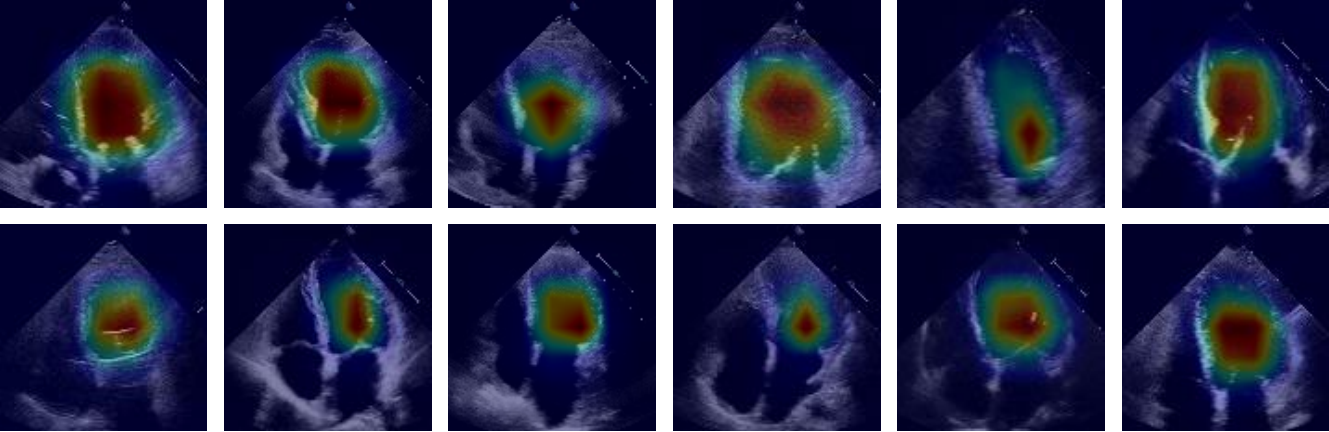}
\caption{The class activation maps of different samples. Each sample represents the first frame of the video clip on the Stanford test set.}\label{cam}
\end{figure*}

\subsection{Robustness Analysis}
In order to verify the robustness of our proposed method, we evaluate it with degraded videos on the Stanford test set. The degraded videos are obtained by randomly replacing the pixels in the original videos with noises whose values are 0. And those degraded videos are not used to train models. As shown in Figure~\ref{noise}, our method is much more robust than the baseline \cite{tran2018closer}, especially in the case of a large noise ratio\footnote{The noise ratio refers to the proportion of the number of noises in the total image pixels.}. In the case of a 50\% noise rate, our method still achieves a respectable performance with a MAE of 4.92, RMSE of 6.52, and $R^2$ of 0.715. Thus, our method is robust to various levels of simulated noise and video degradation.

\subsection{Visualization Analysis}
With our proposed anchor-based classification and regression method, we can utilize the class activation mapping technique \cite{zhou2016learning} for visualization analysis.
The class activation maps reveal the discriminative regions for classification. Due to the reduction in temporal dimension from 32 to 4 before the global average pooling, we can not obtain the class activation map of each frame in the video clip. Therefore, we capture one class activation map for each video clip to reflect the important region in the spatial dimension. As displayed in Figure~\ref{cam}, we can find that our method focuses on the internal area of the left ventricle, which is reasonable.

\bibliographystyle{IEEEbib}

\end{document}